\documentclass{bmvc2k}

\usepackage{marvosym}
\usepackage{ifsym}
\title{RestNet: Boosting Cross-Domain Few-Shot Segmentation with Residual Transformation Network}
 
\addauthor{Xinyang Huang}{hsinyanghuang7@gmail.com}{1}
\addauthor{Chuang Zhu\footnotemark[1]}{czhu@bupt.edu.com}{1}

\addauthor{Wenkai Chen}{wkchen@bupt.edu.com}{1}
\addinstitution{
 School of Artificial Intelligence,\\
 Beijing University of Posts and Telecommunications\\
 Beijing, China
}

\runninghead{Huang, Zhu, Chen}{RestNet: Boosting Cross-Domain Few-Shot}

\usepackage{graphicx}
\usepackage{multirow}
\usepackage{booktabs}
\usepackage{subfigure}
\usepackage{floatrow}
\usepackage{amssymb}
\newfloatcommand{capbtabbox}{table}[][\FBwidth]

\begin{document}

\footnotetext[1]{Corresponding author}
\maketitle

\begin{abstract}
Cross-domain few-shot segmentation (CD-FSS) aims to achieve semantic segmentation in previously unseen domains with a limited number of annotated samples.
Although existing CD-FSS models focus on cross-domain feature transformation, relying exclusively on inter-domain knowledge transfer may lead to the loss of critical intra-domain information.
To this end, we propose a novel residual transformation network (RestNet) that facilitates knowledge transfer while retaining the intra-domain support-query feature information.
Specifically, we propose a Semantic Enhanced Anchor Transform (SEAT) module that maps features to a stable domain-agnostic space using advanced semantics. 
Additionally, an Intra-domain Residual Enhancement (IRE) module is designed to maintain the intra-domain representation of the original discriminant space in the new space.
We also propose a mask prediction strategy based on prototype fusion to help the model gradually learn how to segment.
Our RestNet can transfer cross-domain knowledge from both inter-domain and intra-domain without requiring additional fine-tuning. 
Extensive experiments on ISIC, Chest X-ray, and FSS-1000 show that our RestNet achieves state-of-the-art performance.
Our code is available at \url{https://github.com/bupt-ai-cz/RestNet}.
\end{abstract}
\section{Introduction}
Deep convolutional neural networks (CNNs) have demonstrated remarkable performance in diverse computer vision tasks, including semantic segmentation \cite{long2015fully, zhao2017pyramid, ronneberger2015u} and object detection \cite{dai2017deformable, he2017mask, redmon2016you}.
Although CNNs are effective, their dependence on a large number of labeled data is still a constraint.
To alleviate this dependence, the Few-shot Semantic Segmentation (FSS) \cite{shaban2017one} was proposed to learn the model of segmenting new classes with only a few pixel-level annotations.
In recent years, significant progress has been made in the FSS \cite{zhang2019canet, wang2019panet, siam2019amp, zhang2019pyramid, okazawa2022interclass, liu2020crnet, tian2020prior, min2021hypercorrelation, boudiaf2021few, Liu_2022_BMVC}. 
However, it is still challenging to apply them to cross-domain scenarios. 
To solve this problem, the Cross-Domain Few-Shot Segmentation (CD-FSS) was proposed \cite{lei2022cross} to generalize meta-knowledge from the source domain with sufficient labels to the target domain with limited labels.

\begin{figure}[t]
	\centering
	\subfigure[Previous CD-FFS Method \cite{lei2022cross}]{
		\begin{minipage}{9cm}
  \label{aaa}
                        \includegraphics[width=\textwidth]{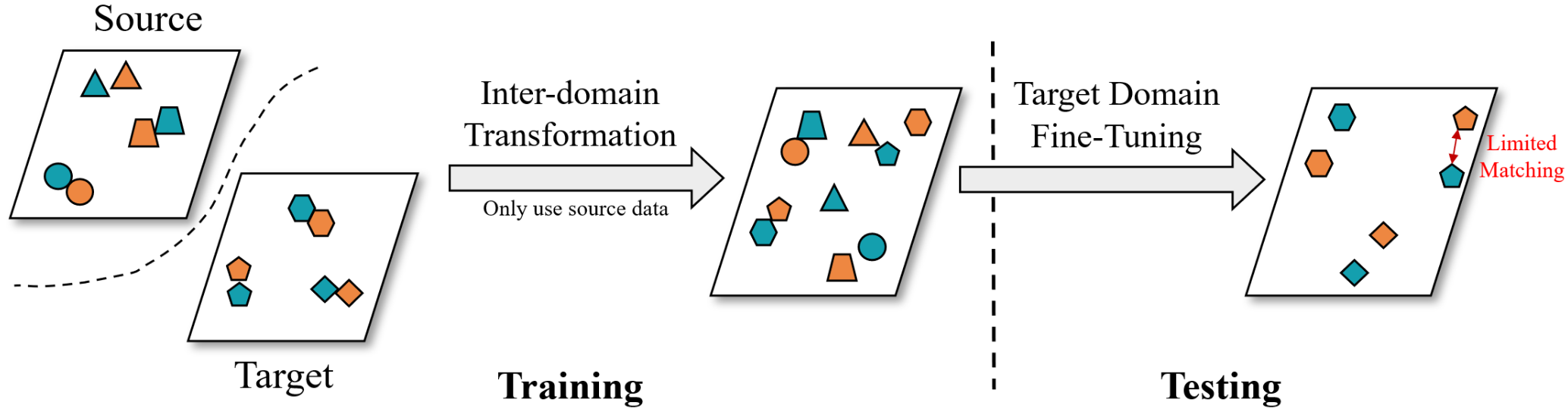} \\
	\vspace{-2.5mm}	
  \end{minipage}
	}\vspace{-3.5mm}
	\subfigure[Our Residual Transformation Network (RestNet)]{
		\begin{minipage}{9.5cm}
  \label{bbb}
			\includegraphics[width=\textwidth]{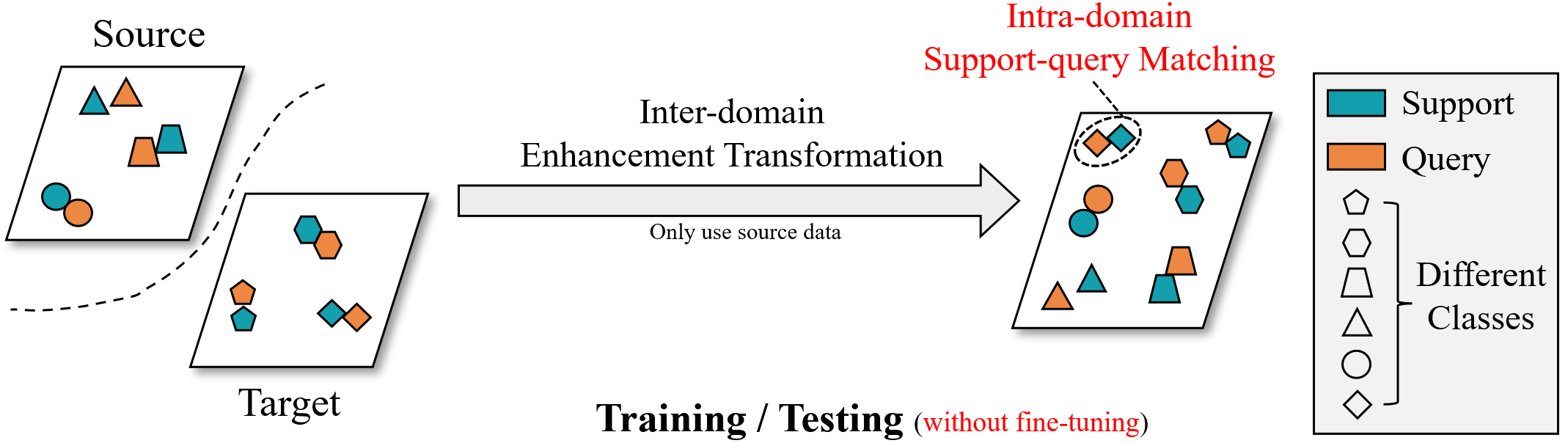} \\
   \vspace{-2.5mm}
		\end{minipage}
	}
 \vspace {-0.5cm}
 \caption{The comparison between the previous cross-domain few-shot segmentation method and our RestNet.
(a) The previous CD-FSS method focused on knowledge transfer using an inter-domain transformation that may lose the intra-domain information, so additional target domain fine-tuning is required.
(b) Our RestNet learns knowledge from both inter-domain and intra-domain.
It performs cross-domain enhancement transformation while preserving the intra-domain matching information.} 
\end{figure}

The CD-FSS problem considers a more realistic scenario: the model cannot access the target data during training, and the data distribution and label space in the test phase are different from those in the training phase. 
To accomplish the CD-FSS, PATNet \cite{lei2022cross} was proposed to perform a linear transformation on the support foreground features and query features, project them into domain-agnostic features, and compute the feature similarity in the new spaces before exporting the query mask.
However, the cross-domain feature mapping by imprecise projection layers often makes the distribution alignment of features aligned in the original space fail in the new space.
In the CD-FSS scenarios, it reflects reduced matching between intra-domain support and query samples, so additional fine-tuning is required in the target domain, as shown in Figure \ref{aaa}.
In addition, due to the fine-grained differences between support and query samples and the presence of support masks, the learning of knowledge may be biased towards the support samples even if they belong to the same class \cite{zha2023boosting}. 
To address these problems, we propose the \textbf{Res}idual \textbf{T}ransformation \textbf{Net}work (RestNet). 
It considers not only inter-domain transfer but also the preservation of intra-domain knowledge, as illustrated in Figure \ref{bbb}.

For the inter-domain, we propose a novel \textbf{S}emantic \textbf{E}nhanced \textbf{A}nchor \textbf{T}ransform (SEAT), which uses the attention to help the model learn advanced semantic features that are then mapped to domain unknown spaces for knowledge migration.
Further, we propose a simple and effective \textbf{I}ntra-domain \textbf{R}esidual \textbf{E}nhancement (IRE) mechanism. 
It associates the information of the original discriminative space to the present domain-agnostic space via residual connection \cite{he2016deep} and helps the model to align the support and query features of the domain-agnostic feature space to enhance the intra-domain knowledge representation. 
The two mechanisms each help the model adapt more comprehensively to cross-domain small-sample tasks from different perspectives. 
Finally, we generate a coarse soft query mask and feed it to the network with the support mask through prototype fusion to obtain the final mask, which can help the model learn how to segment step by step.

In summary, our contributions are summarized as follows:
\begin{itemize}
\vspace{-3.5mm}
\item We propose a Residual Transform network (RestNet) that uses the proposed SEAT and IRE modules to help the model preserve key information in the original domain while performing cross-domain few-shot segmentation.
\vspace{-4mm}
\item We propose a new mask prediction strategy based on the prototype fusion. This strategy helps the proposed RestNet gradually learn how to segment the unseen domain.
\vspace{-8mm}
\item Our method achieves state-of-the-art results on three CD-FSS benchmarks, namely ISIC, Chest X-ray, and FSS-1000. Our RestNet solves the problem of intra-domain knowledge loss under the condition of considering inter-domain knowledge transfer, which provides a new idea for future research in this field.
\end{itemize}

\section{Related Work}
\subsection{Domain Adaptive Segmentation}
Domain adaptive segmentation has led to some achievements. 
The goal of the method is to transfer knowledge learned from a labeled source domain to an unlabeled or weakly labeled target domain. 
The method based on adversarial learning \cite{chen2017no, chen2018road, du2019ssf, huang2022multi} aims to learn domain invariant representations in features. In addition, the design loss function \cite{hsu2021darcnn, liu2021source, wang2021exploring} to constrain the data distribution can also achieve feature alignment. 
These methods operate in settings where the target domain data can be accessed during training to drive model adaption and compensate for domain offsets. 
In contrast, our source and target domain have completely disjoint label spaces and no target data is required during the CD-FSS training.

\subsection{Few-Shot Semantic Segmentation}
Unlike domain adaptive semantic segmentation, the target domain cannot be accessed by the FSS tasks during training. The goal is to segment new semantic objects in the image, with only a few labeled available. 
Current methods mainly focus on the improvement of the meta-learning stage. 
Prototype-based approach \cite{zhang2019canet, wang2019panet, siam2019amp, okazawa2022interclass} is to use methods to extract representative foreground or background prototypes that support data and then use different strategies to interact between different prototypes or between prototypes and query features. 
Relation-based methods \cite{zhang2019pyramid, liu2020crnet, li2021few, tian2020prior, min2021hypercorrelation, Liu_2022_BMVC} also achieved success in the few-shot segmentation. 
HSNet \cite{min2021hypercorrelation} uses multi-scale dense matching to construct hypercorrelation and uses 4D convolution to capture context information.
RePRI \cite{boudiaf2021few} presents a transduction inference for feature extraction on a base class without meta-learning.
However, these methods focus only on segmenting new categories from the same domain. 
Because of the large differences in cross-domain distributions, they cannot be extended to invisible domains.

\section{Method}
\subsection{Problem Setting}
In the problem setting of the CD-FSS, there exists a source domain $(X_s, Y_s)$ and a target domain $(X_t, Y_t)$, where $X$ denotes the distribution of the input data and $Y$ denotes the space of the data label.
The input data distribution of the source domain is different from the target domain, and the label space of the source domain does not intersect with the target domain, i.e., $X_s \neq X_t$ and $Y_s \cap Y_t = \emptyset$.
The model is trained on the source domain and does not have access to the target data.
We place it in a few-shot learning scenario \cite{finn2017meta} for training and inference based on the episode data $(S, Q)$ as same as the previous work \cite{lei2022cross}.
For the few-shot setup, the support set $S = {(I^i_s, M^i_s)}^K_{i = 1}$ contains $K$ image mask pairs, where $I^i_s$ denotes the $i$-th support image and $M^i_s$ denotes the corresponding binary mask. 
Similarly, the query set is defined as $Q = ({I^i_q, M^i_q)}^K_{i = 1}$. 
In the training or meta-training phase, the model obtains the support set $S$ and the query set $I_q$ from a specific class $c$ as input and predicts the mask $M_q$ of the query image. 
In the testing or meta-testing phase, the model performance is evaluated by providing the model with the support set and using the query set of the target domain.
\subsection{Overview}
\begin{figure}[t]
\centering
\includegraphics[scale=0.19]{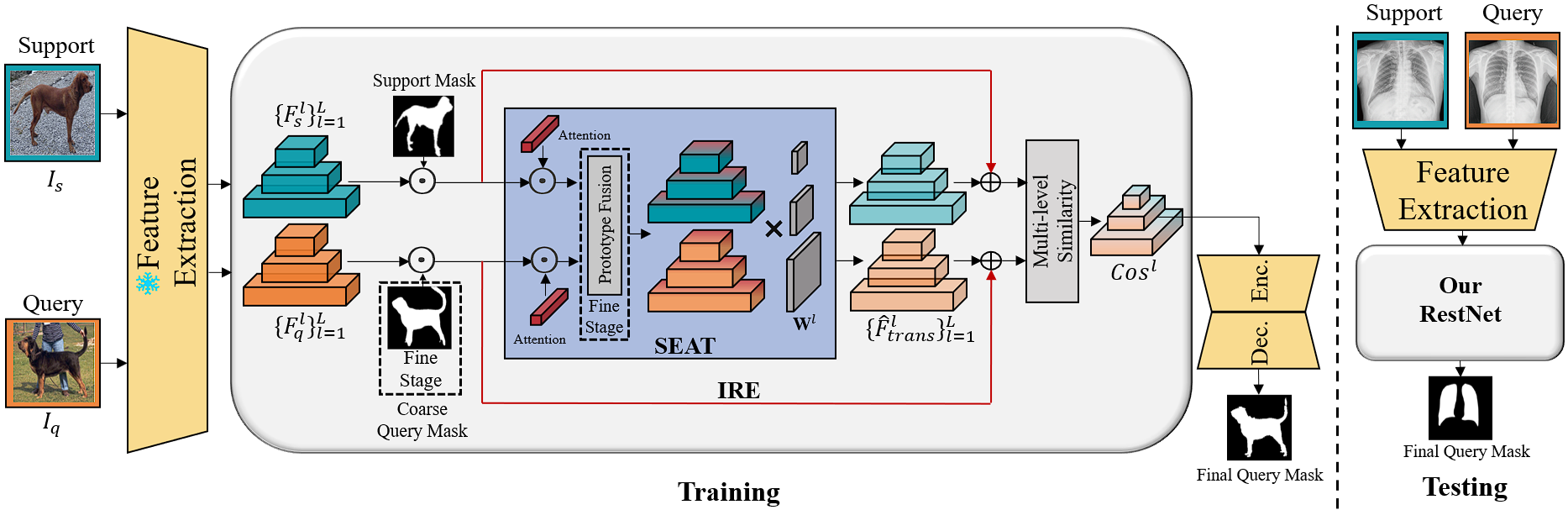}
\caption{The framework of our RestNet.
The model maps support and query features to a new domain-independent space using Semantic Enhanced Anchor Transform (SEAT). The Intra-Domain Residual Enhancement (IRE) module is designed to preserve the matching information of the original feature space in the new space. Next, the similarity between the support and query is calculated and input into the encoder and decoder to generate a rough mask. The final mask is obtained by a prototype fusion mechanism in the fine stage.}
\label{overimg}
\end{figure}

Figure \ref{overimg} illustrates our RestNet, which incorporates two key components to enable rapid CD-FSS adaptation across both inter- and intra-domain: Semantic Enhancement Anchor Transformation (SEAT) and Intra-domain Residual Enhancement (IRE) module.
Given the support set $S = {(I^i_s, M^i_s)}^K_{i = 1}$ and the query image $I_q$, we first derive a multi-level feature map by extracting the features of different layers of the shared backbone weights. 
We map support and query features to a new domain-agnostic space through the semantic enhancement anchor transformation (SEAT).
Then, the intra-domain residual enhancement (IRE) module is designed to preserve the matching information of the original feature space in the new feature space. 
After realignment, we calculate the similarity between the support and query features and input it into the 4D convolution encoder and 2D convolution decoder \cite{min2021hypercorrelation} to generate a coarse query mask.
The coarse mask is fed to the network with the support mask through the prototype fusion mechanism to obtain the final mask.
\subsection{Semantic Enhanced Anchor Transformation}\label{SEAT}
To help the model transfer the cross-domain knowledge, we propose a novel semantic enhancement anchor transformation (SEAT).
The goal is to learn a stable pyramid anchor layer using advanced semantic features derived from a unified attention mechanism to translate features into domain-agnostic features. 
The downstream partitioning module will be easier to predict in such a stable space.

\textbf{Semantic Enhancement. }Before the transformation, the quality of segmentation prediction is highly dependent on the quality of advanced features from the encoder. 
If the encoder fails to provide informative advanced features, it is impossible to obtain useful domain-agnostic features. 
Thus, we use a unified attention mechanism \cite{woo2018cbam} to enhance the support and query feature semantics at intermediate layer $l$:
\begin{eqnarray}
\mathbf{\hat{F}}^l=\sigma(Conv([\operatorname{AvgPool}(\hat{F}^l); \operatorname{MaxPool}(\hat{F}^l)]))\otimes \hat{F}^l,
\end{eqnarray}
where $\hat{F}^l$ denotes the masked feature, $\sigma$ denotes the sigmoid function, $[\mathbf{\cdot};\mathbf{\cdot}]$ represents the concatenation and $\otimes$ represents the element-wise multiplication. 
The unified spatial attention mechanism can share an attention extraction module for the feature layers of different feature channels and support-query samples. 
It can reduce the number of parameters to be learned and help the model learn a unified domain-invariant knowledge between different feature maps from the support and query.

\textbf{Anchor Transformation. }Inspired by the previous work \cite{seo2022task, lei2022cross}, we use the linear transformation matrix as the transformation mapper. 
The matrix can be calculated from the anchor layer parameter matrix and the support prototype. 
Taking 1-way 1-shot as an example, for the intermediate feature layer $\{F^l_s\}_{l=1}^L$ supporting the image, the $l$-th foreground support prototype is as follows:
\begin{eqnarray}
\mathbf{c}^{l}_{s,f}=\frac{\sum_{i}\sum_{j}F^{l,i,j}_s\psi^l(M_{s}^{i,j})}{\sum_{i}\sum_{j}\psi^l(M_{s}^{i,j})},
\end{eqnarray}
where $\psi^l(\cdot)$ denotes the bilinear interpolation, $F^{l,i,j}_s$ and $M_{s}^{i,j}$ represent the pixel values corresponding to the support feature and the mask in row $i$ and column $j$ respectively. 
Similarly, the supporting background prototype $\mathbf{c}^{l}_{s,b}$ can be obtained in the same way. Therefore, given the weight matrix of anchor layer $\mathbf{A}$, the definition of transformation matrix is as follows:
\begin{eqnarray}\label{trans}
\mathbf{W}\mathbf{C}_s=\mathbf{A},
\end{eqnarray}
where $\mathbf{C}_s = \left[\frac{\mathbf{c}_{s,f}}{\|\mathbf{c}_{s,f}\|},\frac{\mathbf{c}_{s,b}}{\|\mathbf{c}_{s,b}\|}\right]$, $\mathbf{A} = \left[\frac{\mathbf{a}_{f}}{\|\mathbf{a}_{f}\|},\frac{\mathbf{a}_{b}}{\|\mathbf{a}_{b}\|}\right]$ and $\mathbf{a}$ is the anchor vector that has the length matching the number of channels in the high-level feature. 
Therefore, we can calculate the transformation matrix $\mathbf{W}$ conveniently. However, since the prototype $\mathbf{C}_s$ is usually a non-square matrix, we can calculate its generalized inverse \cite{ben2003generalized} with $\mathbf{C}^+_{s}=\{\mathbf{C}_{s }^T\mathbf{C}_{s}\}^{-1}\mathbf{C}^{T}_s$. 
Therefore, the transformation matrix of the $l$-th intermediate layer is calculated as
$\textbf{W}^l=\textbf{A}^l\textbf{C}^{l+}_s$.
Similar to \cite{lei2022cross}, we have three anchor layers for low, medium, and high-level features respectively.
Then, we can map the support and query features to stable domain-agnostic space more effectively through this transformation matrix.
\subsection{Intra-domain Residual Enhancement}\label{IRE}
Considering only inter-domain knowledge transfer, the model's ability to perform few-shot tasks in unseen domains is limited. 
This limitation arises from the fact that the features are transformed into a unified space solely through an anchor layer, leading to the loss of crucial intra-domain information. 
This loss is reflected in the reduced matching between support-query features within the same domain. 
To address this issue, we propose an intra-domain residual enhancement module that leverages residual connections to preserve essential information from the original space to help the model perform well in the unseen domain.

Formally, for the transformed feature $\hat{F}^l_{trans}=\mathbf{W}^l \mathbf{\hat{F}}^l$, the definition of intra-domain residual enhancement is as follows:
\begin{eqnarray}
R^l=\hat{F}^l_{trans}\oplus\hat{F}^l,
\end{eqnarray}
where $\oplus$ denotes the residual connection and $\hat{F}^l$ denotes the masked feature.
The residual enhancement module reduces the cross-domain knowledge forgetting caused by anchor layer transformation by introducing support and query information in the original domain.
It does not introduce additional parameters or require additional fine-tuning in the target domain.

For each residual enhancement support-query pair, the cosine similarity is calculated to form a 4D hypercorrelation tensor:
\begin{eqnarray}
\label{SIM}
Cos^l_{i,j}  =  \operatorname{ReLU}\left(\frac{R^l_{s}(i) \cdot R^l_{q}(j)}{\left\|R^l_{s}(i)\right\|\left\|R^l_{q}(j)\right\|}\right),
\end{eqnarray}
where $R^l_{s}(i)$ and $R^l_{q}(j)$ means the $i$-th support 
and $j$-th query residual enhancement feature.
The similarity is fed to the 4D convolution encoder and the 2D convolution decoder to generate the query mask.
The prediction query mask and the ground truth mask are used to calculate the cross-entropy loss, and the model parameters are updated by backpropagation.

\subsection{Mask Prediction Strategy}
Due to the fine-grained difference between the support query samples and the lack of the query mask, the learning of knowledge may also be biased towards the support samples \cite{zha2023boosting}.
At the same time, we can not calculate the query prototype, resulting in the transformation matrix being biased toward the support. 
To solve the above problems, we use the idea of coarse to fine \cite{marr2010vision} to generate a coarse query mask, and then feed it and the support mask to the network through the prototype fusion mechanism to get the final prediction mask.

Specifically, the model first generates a coarse soft mask $\hat{M}_{q}$ for query samples, and the query foreground mask is calculated as follows:
\begin{eqnarray}
\mathbf{\hat{c}}^{l}_{q,f}=\frac{\sum_{i}\sum_{j}F^{l,i,j}_q\psi^l(\hat{M}_{q}^{i,j})}{\sum_{i}\sum_{j}\psi^l(\hat{M}_{q}^{i,j})},
\end{eqnarray}
where $F^{l,i,j}_q$ and $\hat{M}_{q}^{i,j}$ represent the pixel values corresponding to the query feature and the soft mask respectively. 
The background mask for the query can also be calculated in the same way. 
Then we use a simple prototype fusion mechanism to get the final unbiased prototype:
\begin{eqnarray}
\mathbf{c}^{l}_{f}=\alpha\mathbf{c}^{l}_{s,f}+(1-\alpha)\mathbf{\hat{c}}^{l}_{q,f},
\end{eqnarray}
where $\alpha$ is a learnable parameter.
Bring the fused prototype into Equation \ref{trans} to get the exact transformation matrix and go through subsequent modules to get the final query mask. 
This method not only solves the potential phenomenon of supporting sample over-fitting in FSS but also helps the model learn how to segment across domains step by step.
Finally, we optimize the whole model through the cross-entropy loss.

\section{Experiments}
\subsection{Implementation Details}
Following the previous approach, we used PASCAL VOC 2012 \cite{everingham2009pascal} and SBD \cite{hariharan2011semantic} as training domains, and then tested our models on ISIC \cite{tschandl2018ham10000, codella2019skin}, Chest X-ray \cite{candemir2013lung, jaeger2013automatic}, and FSS-1000 \cite{li2020fss}, respectively. 
Each run contains 1200 tasks that contain all datasets except FSS-1000. 
FSS-1000 has 2400 tasks per run \cite{lei2022cross}.
We chose VGG-16 \cite{simonyan2014very} and ResNet-50 \cite{he2016deep} for feature extraction. 
During the training, we kept these weights frozen and selected the feature map as the same as the previous work \cite{lei2022cross}. 
We employed Adam \cite{kingma2014adam} as the optimizer with a learning rate of 1e-3. 
\subsection{Experimental Results}
\begin{figure}[t]
\centering
\includegraphics[scale=0.12]{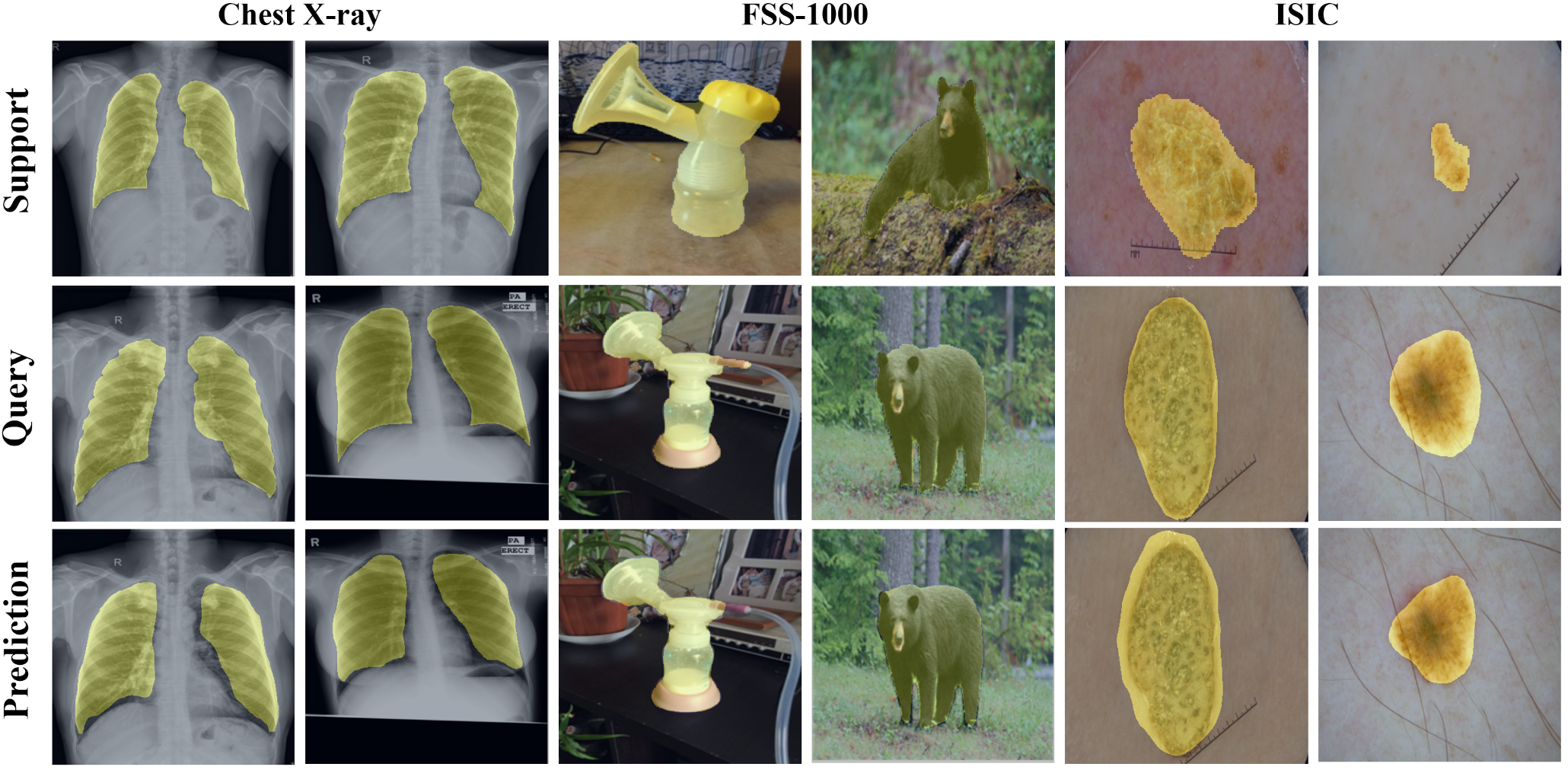}
\caption{Qualitative results of our model for 1-way 1-shot in different CD-FSS datasets. The model is trained using PASCAL VOC. Best color view and zoom in.}
\label{vision}
\end{figure}

As shown in Table \ref{1111111}, our average results at 1-shot and 5-shot show that compared to existing methods, the mIOU of VGG-16 has increased by 3.91\% and 0.59\%, and the mIOU of ResNet-50 has increased by 2.62\% and 1.55\%.
In the case of a large gap between the fields in the source domain dataset, our model achieves SOTA in all results for Chest X-rays.
In ISIC, the mIOU of VGG-16 increased by 3.93\% (1-shot), and the mIOU of ResNet-50 increased by 1.09\% (1-shot).
For FSS-1000, which has a relatively small gap from the source domain dataset, our model surpasses all existing methods and validates its advantages in CD-FSS.
In addition, we show some qualitative results of the proposed method on different datasets in Figure \ref{vision}.
These results prove that our method can improve the generalization ability, which benefits from that our method can learn cross-domain knowledge from different views.
\begin{table}[t]
\centering
\caption{The results of comparison with FSS and CD-FSS methods under 1-way 1-shot and 5-shot settings on the CD-FSS benchmark.
It is noteworthy that all methods are trained in PASCAL VOC and tested on the CD-FSS benchmark.}
\scalebox{0.7}{
\begin{tabular}{c|c|c|c|c|c|c|c|c|c} 
\toprule
\multirow{2}{*}{Methods} & \multirow{2}{*}{Backbone} & \multicolumn{2}{c|}{ISIC} & \multicolumn{2}{c|}{Chest X-ray} & \multicolumn{2}{c|}{FSS-1000} & \multicolumn{2}{c}{Average}  \\ 
\cline{3-10}
                         &                           & 1-shot & 5-shot           & 1-shot & 5-shot                  & 1-shot & 5-shot              & 1-shot & 5-shot              \\ 
\hline
\multicolumn{10}{c}{Few-shot Segmentation Methods}                                                                                                                                \\ 
\hline
AMP \cite{siam2019amp}                 & VGG-16                    & 28.42  & 30.41            & 51.23  & 53.04                   & 57.18  & 59.24               & 45.61  & 47.56               \\
PGNet \cite{zhang2019pyramid}              & ResNet-50                    & 21.86  & 21.25            & 33.95  & 27.96                   & 62.42  & 62.74               & 39.41  & 37.32               \\
PANet \cite{wang2019panet}             & ResNet-50                    & 25.29  & 33.99            & 57.75  & 69.31                   & 69.15  & 71.68               & 50.73  & 58.33               \\
CaNet \cite{zhang2019canet}            & ResNet-50                    & 25.16  & 28.22            & 28.35  & 28.62                   & 70.67  & 72.03               & 41.39  & 42.96               \\
RPMMs \cite{yang2020prototype}               & ResNet-50                    & 18.02  & 20.04            & 30.11  & 30.82                   & 65.12  & 67.06               & 37.75 &39.31
               \\
PFENet \cite{tian2020prior}    & ResNet-50                    & 23.50  & 23.83            & 27.22  & 27.57                   & 70.87  & 70.52               & 40.53 &40.64               \\
RePRI \cite{boudiaf2021few}         & ResNet-50                    & 23.27  & 26.23            & 65.08  & 65.48                   & 70.96  & 74.23               &53.10  & 55.31            \\
HSNet \cite{min2021hypercorrelation} & ResNet-50                    & 31.20  & 35.10            & 51.88  & 54.36                   & 77.53  & 80.99               &53.54 &56.82              \\ 
\hline
\multicolumn{10}{c}{Cross-domain Few-shot Segmentation Methods}                                                                                                                   \\ 
\hline
PATNet \cite{lei2022cross} & VGG-16                    & 33.07  & 45.83            & 57.83  & 60.55                   & 71.60  & 76.17               &54.17
  &60.85             \\
RestNet (Ours)                    &       VGG-16                    &     37.00   &          43.10        &     62.03   &            62.41             &    75.20    &         78.81       &     58.08   &         61.44            \\
PATNet \cite{lei2022cross}                   & ResNet-50                    & 41.16  & \textbf{53.58}            & 66.61  & 70.20              & 78.59  & 81.23               & 62.12 & 68.34             \\
RestNet (Ours)                     &       ResNet-50                    &     \textbf{42.25}   &       51.10           &     \textbf{70.43}   &                \textbf{73.69}         &    \textbf{81.53}    &        \textbf{84.89}             &      \textbf{64.74} &      \textbf{69.89}       \\
\bottomrule
\end{tabular}
}
\vspace {-0.5cm}
\label{1111111}
\end{table}
\subsection{Ablation Study}
\subsubsection{Component Analysis}
Our method mainly includes three parts, namely the Semantic Enhanced Anchor Transformation (SEAT) module and the Residual Enhancement (IRE) module, and Mask Prediction Strategy (MPS).
We validated the effectiveness of each component and presented the results in Table \ref{tab:tb2}.
It can be concluded that SEAT and IRE have brought significant improvements to the model from different perspectives, and MPS is also indispensable.

\subsubsection{Effect of Different Attention}
We mentioned a unified attention mechanism in Section \ref{SEAT}.
In Table \ref{tab:tb3}, we calculated its results on the FSS-1000 with different attention mechanisms (i.e., support and query non-shared attention mechanism modules).
The results show that a unified attention mechanism can not only help the model reduce learnable parameters but also help the model learn unified support query attention information to achieve better segmentation results.

\begin{table*}[ht]
\begin{floatrow}
\capbtabbox{
 \scalebox{0.6}{
\begin{tabular}{cccc|c}
\toprule
Baseline & SEAT & IRE & MPS & 1-shot mIOU \\
\midrule
  \checkmark &  &     &     & 77.53        \\
 \checkmark &\checkmark    &     &     & 78.21        \\
 \checkmark &\checkmark    &  \checkmark   &     & 81.03        \\
 \checkmark &\checkmark    &  \checkmark   &  \checkmark   & \textbf{81.53}   \\
\bottomrule
\end{tabular}
}
}{
 \caption{Ablation study of key modules on FSS-1000.}
 \label{tab:tb2}
}
\capbtabbox{
 \scalebox{0.7}{
\begin{tabular}{ccc|c}
\toprule
Method & 1-shot mIOU \\
\midrule
Different Attention  & 81.69        \\
Unified Attention   & \textbf{82.03}        \\
\bottomrule
\end{tabular}
}
}{
 \caption{Ablation study of different attention on FSS-1000.}
 \label{tab:tb3}
}
\end{floatrow}
\end{table*}

\subsection{Additional Analysis}
\subsubsection{Visualization of Intra-domain Knowledge}
\begin{figure*}
	\centering
	\subfigure[Activate Matching]{
		\includegraphics[width=0.36\linewidth]{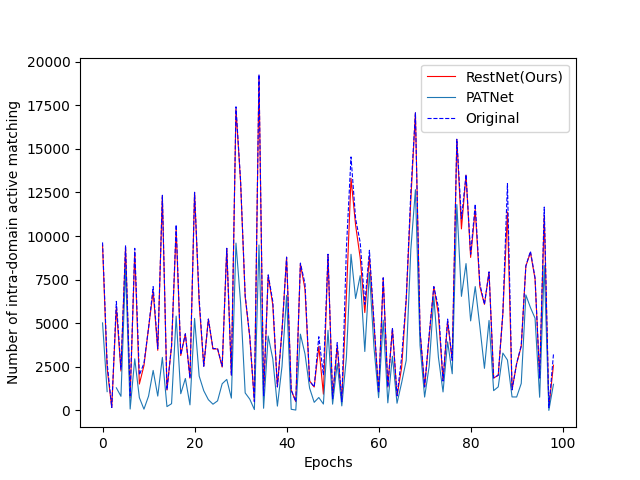}
  \label{AM1}}\hspace{-3mm}
	\subfigure[PATNet \cite{lei2022cross}]{
		\includegraphics[width=0.31\linewidth]{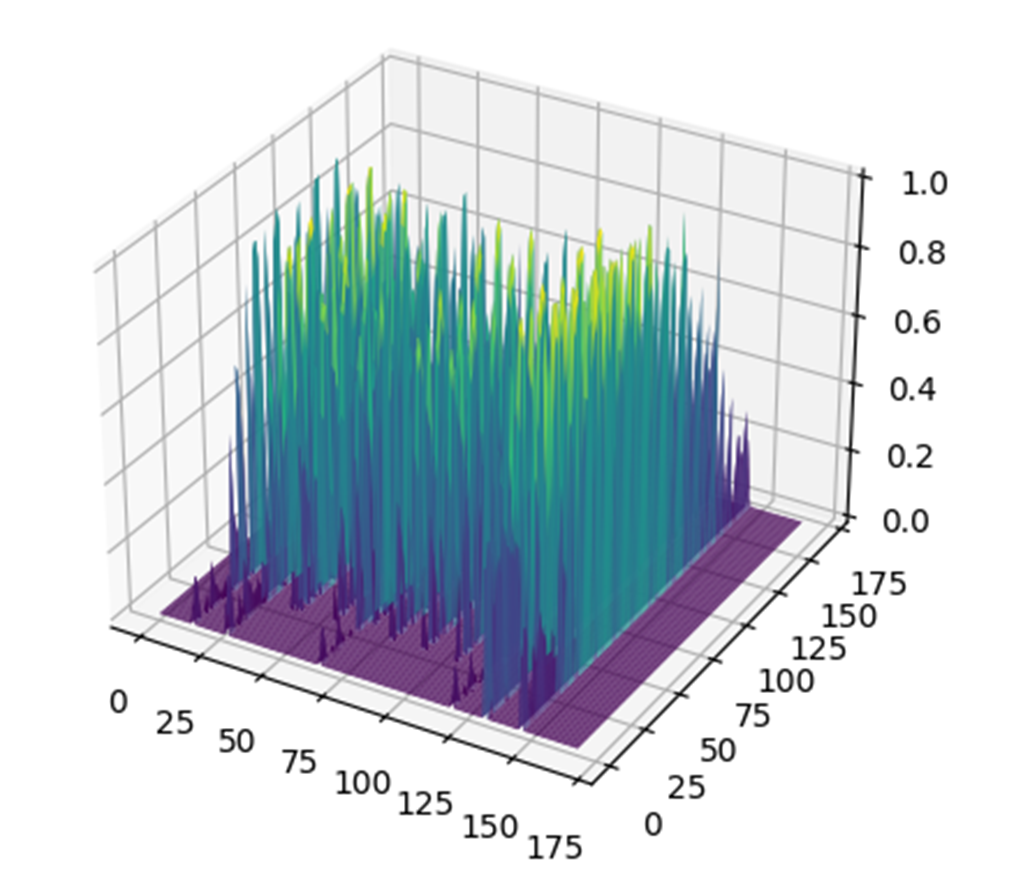}\label{AM2}}
	\subfigure[RestNet (Ours)]{
		\includegraphics[width=0.28\linewidth]{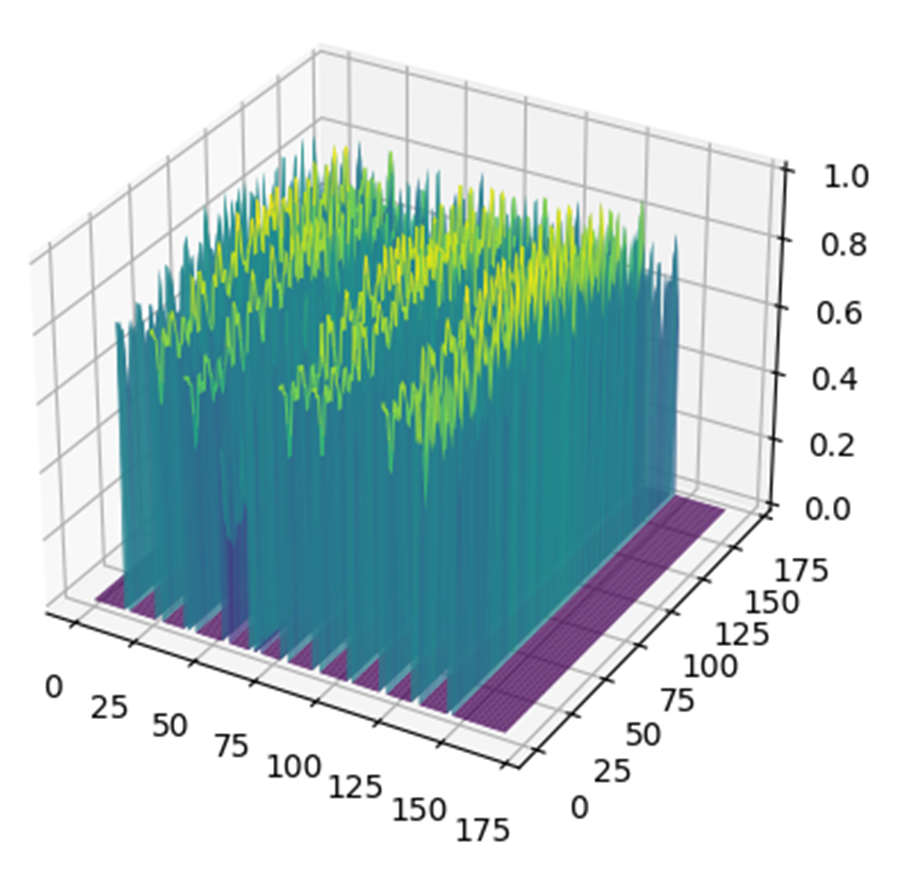}\label{AM3}}
  \vspace {-0.5cm}
	\caption{Visualization of the Intra-domain support-query activate matching.}
	\label{fig2}
\end{figure*}
In Section \ref{IRE}, we mentioned the importance of intra-domain matching similarity.
To quantify this attribute, for the same domain support and query samples, we calculate the support-query similarity in Equation \ref{SIM} and count the pixel pairs in each similarity whose value is greater than zero.
We call these pixel pairs \textit{intra-domain active matching}.
The Intra-domain active matching reflects the knowledge learning between support and query samples from the same domain, which can well represent the retention of intra-domain knowledge by the model after cross-domain feature projection in the CD-FSS.

Specifically, we visualized the active matching of our method and existing methods \cite{lei2022cross} during the training process, as shown in Figure \ref{AM1}, where \textit{Original} denotes no feature transformation.
Compared to existing methods, our method can well preserve the intra-domain knowledge in the original space and reduce the loss caused by cross-domain transformation.
Further, we visualize the support-query similarity for the 30th epoch. 
For ease of visualization, we reshape the similarity from $\mathbb{R}^{\in H\times W\times H\times W}$ to $\mathbb{R}^{\in HW\times HW}$, as shown in Figure \ref{AM2} and Figure \ref{AM3}.
It can be shown that our method can activate more intra-domain matching, helping the model utilize more intra-domain support-query information.

\subsubsection{Comparison of Model Parameter Quantities}
In Table \ref{param}, we compare the number of parameters with those of existing FSS methods and CD-FSS methods.
We calculate the number of additional parameters relative to the backbone.
The results show that our method only introduces a very small number of parameters but has additional performance improvements compared with the recent CD-FSS method.
\begin{table}[t]
\caption{Comparison of the number of additional parameters of different methods.}
\centering
\scalebox{0.65}{
\begin{tabular}{c|c|c|c}
\toprule
Method        & HSNet \cite{min2021hypercorrelation}  & PATNet \cite{lei2022cross}& RestNet (Ours)  \\
\midrule
Additional Parameters (M) & 2.5740 M  & 2.5809 M  & 2.5812 M      \\
\bottomrule
\end{tabular}
}
\label{param}
\end{table}
\section{Conclusion}
In this work, we propose a residual transform network (RestNet) to solve the cross-domain few-shot segmentation (CD-FSS). 
It is to comprehensively help the model to transfer knowledge with few samples from both inter-domain and intra-domain perspectives. 
To achieve this, we propose a Semantic Enhanced Anchor Transformation (SEAT) module to help the model learn domain-independent features using advanced semantic features. 
In addition, an intra-domain residual enhancement (IRE) module is also involved to help the model enhance intra-domain information while transferring knowledge between domains.
Finally, we use the mask prediction strategy based on prototype fusion to help the model gradually learn how to segment the unseen domain. 
On the three CD-FSS benchmarks, several experiments have proved that our RestNet achieves state-of-the-art performance.

\section*{Acknowledgement}
This work was supported by the National Key R\&D Program of China (2021ZD0109800), by the National Natural Science Foundation of China (81972248).

\bibliography{egbib}

\end{document}